\documentclass[letterpaper]{article} 
\usepackage{aaai24}  
\usepackage{times}  
\usepackage{helvet}  
\usepackage{courier}  
\usepackage[hyphens]{url}  
\usepackage{graphicx} 
\usepackage{natbib}  
\usepackage{caption} 
\usepackage{algorithm}
\usepackage{algorithmic}
\usepackage{newfloat}
\usepackage{listings}
\DeclareCaptionStyle{ruled}{labelfont=normalfont,labelsep=colon,strut=off} 
\floatstyle{ruled}
\newfloat{listing}{tb}{lst}{}
\floatname{listing}{Listing}
\usepackage{lmodern}
\usepackage[frozencache,cachedir=.]{minted}
\usepackage{cite}
\usepackage[many]{tcolorbox}
\usepackage{xcolor}
\usepackage{varwidth}
\usepackage{environ}
\usepackage{xparse}
\usepackage{graphicx}
\usepackage{multirow}
\usepackage{tabularx}
\usepackage{mdframed}
\usepackage{mathtools}
\usepackage{caption}
\usepackage{makecell}
\usepackage{pifont}
\usepackage{nameref}
\newcommand{\xmark}{\ding{55}}%

\title{GPT-4V Takes the Wheel: Promises and Challenges for Pedestrian Behavior Prediction}
\author{
    Jia Huang$^*$,
    Peng Jiang$^*$,
    Alvika Gautam,
    Srikanth Saripalli
}
\affiliations{
    Texas A\&M University\\

    {\{jia.huang,maskjp,alvikag,ssaripalli\}@tamu.edu
}
}

\usepackage{bibentry}
\newlength{\bubblesep}
\newlength{\bubblewidth}
\AtBeginDocument{\setlength{\bubblewidth}{.43\textwidth}}
\definecolor{bubblegreen}{RGB}{255,255,224}
\definecolor{bubblegray}{RGB}{240,248,255}

\newcommand{\bubble}[4]{%
  \tcbox[
    on line,
    arc=1mm,
    colback=#1,
    colframe=#1,
    left=0.5 pt,
    right=0.5 pt,
    top=0.5 pt,
    bottom=0.5 pt, 
    before={\vspace{-0.08cm}}, 
    after={\vspace{-0.08cm}},
    #2,
  ]{\color{#3}\begin{varwidth}{\bubblewidth}\tiny#4\end{varwidth}}%
}

\ExplSyntaxOn
\seq_new:N \l__ooker_bubbles_seq
\tl_new:N \l__ooker_bubbles_first_tl
\tl_new:N \l__ooker_bubbles_last_tl
\NewEnviron{rightbubbles}
 {
  \begin{flushright}
  \sffamily
  \seq_set_split:NnV \l__ooker_bubbles_seq { \par } \BODY
  \int_compare:nTF { \seq_count:N \l__ooker_bubbles_seq < 2 }
   {
    \bubble{bubblegreen}{rounded~corners}{black}{ \BODY}\par
   }
   {
    \seq_pop_left:NN \l__ooker_bubbles_seq \l__ooker_bubbles_first_tl
    \seq_pop_right:NN \l__ooker_bubbles_seq \l__ooker_bubbles_last_tl
    \bubble{bubblegreen}{sharp~corners=southeast}{black}{\l__ooker_bubbles_first_tl}
    \par\nointerlineskip
    \addvspace{\bubblesep}
    \seq_map_inline:Nn \l__ooker_bubbles_seq
     {
      \bubble{bubblegreen}{sharp~corners=east}{black}{##1}
      \par\nointerlineskip
      \addvspace{\bubblesep}
     }
    \bubble{bubblegreen}{sharp~corners=northeast}{black}{\l__ooker_bubbles_last_tl}
    \par
   }
   \end{flushright}
 }
\NewEnviron{leftbubbles}
 {
  \begin{flushleft}
  \sffamily
  \seq_set_split:NnV \l__ooker_bubbles_seq { \par } \BODY
  \int_compare:nTF { \seq_count:N \l__ooker_bubbles_seq < 2 }
   {
    \bubble{bubblegray}{rounded~corners}{black}{\BODY}\par
   }
   {
    \seq_pop_left:NN \l__ooker_bubbles_seq \l__ooker_bubbles_first_tl
    \seq_pop_right:NN \l__ooker_bubbles_seq \l__ooker_bubbles_last_tl
    \bubble{bubblegray}{sharp~corners=southwest}{black}{\l__ooker_bubbles_first_tl}
    \par\nointerlineskip
    \addvspace{\bubblesep}
    \seq_map_inline:Nn \l__ooker_bubbles_seq
     {
      \bubble{bubblegray}{sharp~corners=west}{black}{##1}
      \par\nointerlineskip
      \addvspace{\bubblesep}
     }
    \bubble{bubblegray}{sharp~corners=northwest}{black}{\l__ooker_bubbles_last_tl}\par
   }
  \end{flushleft}
 }
\ExplSyntaxOff

\newtcolorbox{chatbox}[1][]{
    enhanced,
    breakable,
    frame hidden,
    colback = white!10,
    left=1 pt,
    right=1 pt,
    top=1 pt,
    bottom=3 pt,
    overlay unbroken= {
        \draw[line width=0.5mm, gray, rounded corners]
        (frame.north west) rectangle (frame.south east);
    },
    overlay broken = {
        \draw[line width=0.5mm, gray, rounded corners]
        (frame.north west) rectangle (frame.south east);
    },
    coltitle=black,
    colbacktitle=orange!25,
    #1 
}
\begin{document}

\maketitle
\def\thefootnote{*}\footnotetext{These authors contributed equally to this work.}

\begin{abstract}

Predicting pedestrian behavior is the key to ensure safety and reliability of autonomous vehicles. While deep learning methods have been promising by learning from annotated video frame sequences, they often fail to fully grasp the dynamic interactions between pedestrians and traffic, crucial for accurate predictions. These models also lack nuanced common sense reasoning. Moreover, the manual annotation of datasets for these models is expensive and challenging to adapt to new situations. The advent of Vision Language Models (VLMs) introduces promising alternatives to these issues, thanks to their advanced visual and causal reasoning skills. To our knowledge, this research is the first to conduct both quantitative and qualitative evaluations of VLMs in the context of pedestrian behavior prediction for autonomous driving. We evaluate GPT-4V(ision) on publicly available pedestrian datasets: JAAD and WiDEVIEW. Our quantitative analysis focuses on GPT-4V's ability to predict pedestrian behavior in current and future frames. The model achieves a 57\% accuracy in a zero-shot manner, which, while impressive, is still behind the state-of-the-art domain-specific models (70\%) in predicting pedestrian crossing actions. Qualitatively, GPT-4V shows an impressive ability to process and interpret complex traffic scenarios, differentiate between various pedestrian behaviors, and detect and analyze groups. However, it faces challenges, such as difficulty in detecting smaller pedestrians and assessing the relative motion between pedestrians and the ego vehicle. 

\end{abstract}

\section{Introduction}
To ensure safe autonomous driving and make timely maneuvering decisions, it is crucial for Autonomous Vehicles (AVs) to have the ability to recognize and anticipate pedestrians' behaviors effectively, especially in urban environments with complex vehicle-pedestrian interactions. Pedestrian behavior prediction is a challenging task due to diverse factors influencing pedestrians' behaviors, which include their individual characteristics like demographics and gaits, social interactions with other road users (e.g. whether walking in a group or alone), responses to environmental factors such as but not limited to road width, traffic lights etc. \cite{Rasouli_survey2020}. As high-level information of behavior is not directly observable and cannot be estimated by simply using the pedestrian's trajectories, pedestrian crossing intention prediction requires a holistic comprehension of the context, scene, pedestrian behavioral attributes, and meticulous inference from past actions \cite{SHARMA2022120,zhang2023survey}. 

Most of the recent works treat pedestrian crossing intention prediction as a binary classification with crossing or not-crossing (C/NC) action, while some other studies predict multi-classification with several different action types such as crossing, stopping, bending, and starting in \cite{fang2018pedestrian}. These methods make use of one or more feature inputs, such as pedestrian poses \cite{fang2018pedestrian,pedestrian_pose} or skeleton \cite{HMM}, past trajectories and velocities through ground truth annotations or real time tracking algorithms\cite{huangLearningPedestrianActions2023, saleh2019real}, local context\cite{JAAD_cite1}, semantic maps\cite{Rasoulihybrid2022}, ego vehicle dynamics\cite{Rasoulihybrid2022,kotseruba2021benchmark}, etc. These features are concatenated and then fed into sequential models like RNN \cite{yang2022rnnST}, LSTM and transformer-based models \cite{huangLearningPedestrianActions2023,zhou2023transformer,sui2021tf}, or non-sequential models like CNN \cite{saleh2019real} and GNN-based models \cite{razali2021pedestrian,chenGCN2021,yau2021graph} to capture temporal and spatial information \cite{spatiotemporal}.

Although these vision-based methods show promising results, they exhibit weaknesses in accurately perceiving objects on real-life data, and have difficulty in interpreting the behavioral intentions of surrounding traffic participants in complex and rapidly dynamic environments. Moreover, they struggle to distill driving-related knowledge from data for nuanced scenario understanding and effective causal reasoning, leading to potential safety concerns, and limiting the path toward more advanced autonomous driving.

The emergence of VLMs provide potential solutions for the inherent limitations of current autonomous driving tasks. While Large Language Models (LLMs) provide human-like understanding and reasoning capabilities for decision making,  VLMs \cite{zhu2023minigpt,peng2023kosmos} including GPT-4 \cite{OpenAI2023GPT4TR} further expand LLMs' capabilities through the inclusion and reasoning of image inputs, thus extending their functionality beyond strictly text-based interaction \cite{fuDriveHumanRethinking2023,cuiDriveLLMChartingPath2023,xu2023drivegpt4}.

Recently, the advent of the cutting edge VLM, GPT-4V \cite{gallagherGPT4VisionComplete2023} has expanded the horizons for research and development. An evaluation of GPT-4V in autonomous driving \cite{wenRoadGPT4VIsion2023} scenarios demonstrates its ability to understand and reason about driving scenes and make decisions as a driver, including traffic light recognition and vision grounding. Although similar works touch upon pedestrian-vehicle interactions in temporal sequences reasoning of autonomous driving tasks, to the best of our knowledge, none of them has focused on the ability of GPT-4V for pedestrian behavior prediction both quantitatively and qualitatively.

Towards this, we evaluate the performance of GPT-4V on the most widely used pedestrian behavior dataset JAAD \cite{JAAD_cite1} along with our own dataset WiDEViEW \cite{huangWiDEVIEWUltraWideBandVision2023}, and provide a thorough assessment of the extent to which VLM can bring to the landscape of human-centric autonomous driving and VLM's ability to decipher and reason about pedestrian and road users' behaviour and eventually their role in safer AV-human interactions.

\section{Evaluation Method}
\subsection{Datasets}
JAAD \cite{JAAD_cite1} is one of the publicly available datasets specifically designed for pedestrian crossing action prediction. In contrast, WiDEVIEW \cite{huangWiDEVIEWUltraWideBandVision2023} serves as a comprehensive multi-modal dataset, for deciphering pedestrian-vehicle interactions in urban environments. Our quantitative evaluation relies only on JAAD, while selective interesting scenarios from both datasets are discussed for qualitative analysis. Table \ref{tab:dataset} shows the main features of the datasets used for evaluation.
\begin{table}[h]
\noindent
\scriptsize
\begin{tabular}{|p{1.3cm}|p{3.4cm}|p{2.3cm}|}
\hline
Dataset & Features & Annotations \\  \hline 
JAAD & \makecell[l]{Image-centric dataset,\\Ped bboxes tracks, \\ Behavioural and attri-
\\butes annotations.} & \makecell[l]{Ped bboxes:\\ 378,643 \\
Behavioral tags:\\ 686 ped} \\ \hline 
WiDEViEW & \makecell[l]{Multimodal dataset\\ $(LiDAR, image, UWB)$,\\
Various ped-veh interactions,\\
Ped bboxes tracks and \\UWB tag ID annotations\\
}
 & \makecell[l]{Ped bboxes: \\8,368 manual,\\
33,239 interpolated\\} \\ \hline
\end{tabular}
\caption{Key features of the datasets used for GPT-4V evaluation. Bboxes, ped and veh denote bounding boxes, pedestrians and vehicles respectively. }
\label{tab:dataset}
\end{table}
\subsection{Quantitative Experiment Design}
For quantitative experiments, we utilize the OpenAI Python API to perform a batch evaluation on JAAD dataset. We task the model with inferring information from $n$ (\texttt{num\_frames}) past frames and predicting pedestrian behavior for $m$ (\texttt{prediction\_num}) future frames. Key behaviors we focus on, annotated in the JAAD dataset, include $crossing$, $action(walking/standing)$ and $looking$. The JAAD dataset videos are captured at a frequency of 30 FPS. Our preliminary experiments indicate that the model struggles to discern differences between frames that are too closely spaced due to subtle motion changes. While an ideal experiment would vary \texttt{skip\_num} (the number of frames to skip), API constraints limit us to perform batch analysis with different \texttt{skip\_num}. Consequently, we set \texttt{num\_frames} to 10 and \texttt{prediction\_num} to 5.

Our prompts are constructed based on best practices (see \cite{BestPracticesPrompt}) and iterative trial-and-error. The system message is designed to contextualize the AI’s role as an autonomous vehicle with a front camera that interacts with pedestrians and is capable of outputting JSON files. The prompt is as follows:
\begin{minted}[frame=single, framesep=1mm, linenos=false, xleftmargin=2pt, tabsize=2, fontsize=\scriptsize, breaklines]{python}
"You are an autonomous vehicle that uses front-camera images to interact with pedestrians and also a helpful assistant designed to output JSON."
\end{minted}
Subsequently, we input image details along with the definitions of the targeted behaviors for analysis. Rather than requesting predictions for all pedestrians in the image at the same time, we focus our inquiry on a single pedestrian at a time, distinguished by a red bounding box, which aligns with standard practices in pedestrian behavior prediction research. Since the sizes of some pedestrians are too small even for humans to accurately detect, we filter out some pedestrian samples based on their bounding boxes sizes. As a result, we are left with 194 sequences for evaluation.

The images are entered sequentially and initially we used the definitions of pedestrian behavior provided in the JAAD dataset. However, these definitions were not ideal for the model's understanding and task performance based on initial tests. Consequently, we turned to ChatGPT for help in refining these definitions. This modification leads to a noticeable improvement in the model's performance, as the revised definitions are more conducive to its comprehension and predictive capabilities compared to the original ones.
\begin{minted}[frame=single, framesep=1mm, linenos=false, xleftmargin=2pt, tabsize=2, fontsize=\scriptsize, breaklines]{python}
f"""These are {num_frames} ego-vehicle front-camera view images that you can see behind the wheel.\
Here are the definitions of the pedestrian behaviors we are interested in: \
* Cross: Not-crossing: The pedestrian is not crossing the road .\
Crossing: The pedestrian is actively crossing the road and their path intersects with that of the ego-vehicle.\
Crossing-irrelevant: The pedestrian is crossing the road, but their path does not intersect with the ego-vehicle's path, hence it is irrelevant to the vehicle's immediate trajectory.\
* Action: Whether the pedestrian is `walking` or `standing` \
* Look: This term describes the pedestrian's attention in relation to the ego-vehicle, indicating if the pedestrian is 'looking' towards or 'not-looking' (not-facing) away from the direction of the ego-vehicle. """
\end{minted}
Our approach to inference and prediction involves posing specific questions to the model. Initially, these questions were directly tied to the three key behaviors we were investigating. However, we encountered a challenge: the model struggled to distinguish between 'crossing' and 'walking along a sidewalk'. To mitigate this issue, we introduced an additional question, which was proved to be somewhat effective, enhancing the model's ability to differentiate between these two behaviors.
\begin{minted}[frame=single, framesep=1mm, linenos=false, xleftmargin=2pt, tabsize=2, fontsize=\scriptsize, breaklines]{python}
f"""By answering the following questions, can you describe the current {num_frames} frames and predict the next {prediction_num} frame behavior of the pedestrian within the red box separately?                
Questions: \
* Can you drive forward very fast without hitting the pedestrian within the red box ? \
* Is the pedestrian crossing in front of our car? \
* What's the action (standing/walking) of the pedestrian? \
* Is the pedestrian looking towards the direction of the ego-vehicle (not-looking/looking)?` """
\end{minted} 
In the final step of our procedure, we direct the model to present its responses in a predefined JSON format. This structured output is essential for consistent analysis and interpretation of the results.
We consolidate all the prompts into a single text-based input, followed by a series of image prompts. Given that GPT is a generative model, its outputs can vary with each iteration. To enhance determinism in our experiments, we configure internal parameters of the GPT model, setting 'temperature' to 0 and 'seed' to 0. Additionally, to ensure consistency and reliability, we repeat the same prompt sequence five times for each experiment. 

\subsection{Qualitative Experiment Design}

Our qualitative experiments with ChatGPT supplement the insights from quantitative batch experiments, enhancing our understanding of GPT-4V's capabilities in real-world scenarios. We expose GPT-4V to a variety of images from both datasets to assess its initial responses, then engage in interactive dialogues to probe deeper into its initial observations and analyze the model's logic. We document the process to profile the model's performance, capturing both accurate interpretations and misjudgments.

\begin{table*}[!h]
\scriptsize	
\centering
\begin{tabular}{|c|c|c|cc|ccccc}
\hline
\multirow{2}{*}{Models} & \multirow{2}{*}{Year} & \multirow{2}{*}{Model Variants} & \multicolumn{2}{c|}{Input}                       & \multicolumn{5}{c}{JAAD}                                                                                                                                \\ \cline{4-10} 
                        &                       &                                 & \multicolumn{1}{c|}{Use Frames} & Extra Info     & \multicolumn{1}{c|}{ACC}  & \multicolumn{1}{c|}{AUC}  & \multicolumn{1}{c|}{$\mathrm{F} 1$} & \multicolumn{1}{c|}{$P$}  & \multicolumn{1}{c|}{$\mathrm{R}$} \\ \hline
PCPA                    & 2021                  & 3D CNN+RNN+Attention            & \multicolumn{1}{c|}{16}         & \xmark            & \multicolumn{1}{c|}{0.58} & \multicolumn{1}{c|}{0.50} & \multicolumn{1}{c|}{0.71}           & \multicolumn{1}{c|}{\textbackslash}    & \multicolumn{1}{c|}{\textbackslash}            \\ \hline
TrouSPI-Net             & 2021                  & GRU+Attention                   & \multicolumn{1}{c|}{16}         & \xmark            & \multicolumn{1}{c|}{0.64} & \multicolumn{1}{c|}{0.56} & \multicolumn{1}{c|}{0.76}           & \multicolumn{1}{c|}{0.66} & \multicolumn{1}{c|}{0.91}         \\ \hline
IntFormer               & 2021                  & Transformer                     & \multicolumn{1}{c|}{16}         & \xmark            & \multicolumn{1}{c|}{0.59} & \multicolumn{1}{c|}{0.54} & \multicolumn{1}{c|}{0.69}           & \multicolumn{1}{c|}{\textbackslash}    & \multicolumn{1}{c|}{\textbackslash}            \\ \hline
ST CrossingPose         & 2022                  & Graph CNN                       & \multicolumn{1}{c|}{16}         & \xmark            & \multicolumn{1}{c|}{0.63} & \multicolumn{1}{c|}{0.56} & \multicolumn{1}{c|}{0.74}           & \multicolumn{1}{c|}{0.66} & \multicolumn{1}{c|}{0.83}         \\ \hline
FFSTP                   & 2022                  & GRU+Attention                   & \multicolumn{1}{c|}{16}         & $Seg$            & \multicolumn{1}{c|}{0.62} & \multicolumn{1}{c|}{0.54} & \multicolumn{1}{c|}{0.74}           & \multicolumn{1}{c|}{0.65} & \multicolumn{1}{c|}{0.85}         \\ \hline
Pedestrian Graph +      & 2022                  & Graph CNN+Attention             & \multicolumn{1}{c|}{32}         & $Seg$, $P_{3 D}$ & \multicolumn{1}{c|}{\textcolor{blue}{\textbf{0.70}}} & \multicolumn{1}{c|}{\textcolor{blue}{\textbf{0.70}}} & \multicolumn{1}{c|}{0.76}           & \multicolumn{1}{c|}{0.77} & \multicolumn{1}{c|}{0.75}         \\ \hline
PIT-Block(a)            & 2022                  & Transformer                     & \multicolumn{1}{c|}{16}         & \xmark            & \multicolumn{1}{c|}{\textcolor{blue}{\textbf{0.70}}} & \multicolumn{1}{c|}{0.65} & \multicolumn{1}{c|}{\textcolor{blue}{\textbf{0.81}}}           & \multicolumn{1}{c|}{0.71} & \multicolumn{1}{c|}{\textcolor{blue}{\textbf{0.93}}}         \\ \hline
PIT-Block(d)            & 2022                  & Transformer                     & \multicolumn{1}{c|}{16}         & \xmark            & \multicolumn{1}{c|}{\textcolor{blue}{\textbf{0.70}}} & \multicolumn{1}{c|}{0.69} & \multicolumn{1}{c|}{0.76}           & \multicolumn{1}{c|}{0.79} & \multicolumn{1}{c|}{0.74}         \\ \hline
\textcolor{red}{\textbf{GPT-4V} } & \textcolor{red}{\textbf{2023}} & \textcolor{red}{\textbf{Transformer} } & \multicolumn{1}{c|}{\textcolor{red}{\textbf{10}}} & text prompt & \multicolumn{1}{c|}{\textcolor{red}{\textbf{0.57}}} & \multicolumn{1}{c|}{\textcolor{red}{\textbf{0.61}}} & \multicolumn{1}{c|}{\textcolor{red}{\textbf{0.65}}} & \multicolumn{1}{c|}{\textcolor{blue}{\textbf{0.82}}} & \multicolumn{1}{c|}{\textcolor{red}{\textbf{0.54}}} \\ \hline
\textcolor{red}{\textbf{GPT-4V Skip} } & \textcolor{red}{\textbf{2023}} & \textcolor{red}{\textbf{Transformer} } & \multicolumn{1}{c|}{\textcolor{red}{\textbf{10}}} & text prompt & \multicolumn{1}{c|}{\textcolor{red}{\textbf{0.55}}} & \multicolumn{1}{c|}{\textcolor{red}{\textbf{0.59}}} & \multicolumn{1}{c|}{\textcolor{red}{\textbf{0.64}}} & \multicolumn{1}{c|}{\textcolor{blue}{\textbf{0.81}}} & \multicolumn{1}{c|}{\textcolor{red}{\textbf{0.53}}} \\ \hline 
\end{tabular}
\caption{Performance comparison with state-of-the-art methods from PIT Paper \cite{zhou2023transformer}. Blue is the best. }
\label{tab:benchmark}
\end{table*}
\section{Results}
\subsection{Quantitative Experiments}
\subsubsection{Evaluation Metrics}
In this study, we approach the pedestrian behavior prediction as a standard classification task. To achieve a comprehensive view of the model's prediction accuracy and reliability, we employ several key standard metrics: accuracy, precision, recall, F1 score, and ROC AUC (Area
Under the Curve).
To quantitatively evaluate the model's inherent randomness in output, we utilize entropy to measure the differences across multiple predictions generated by the model.
\begin{table}[h]
\scriptsize		
\centering
\begin{tabular}{|l|l|l|l|l|l|l|} \hline
Time & Behavior & ACC & AUC & $\mathrm{F} 1$ & $P$ & $\mathrm{R}$ \\ \hline
\multirow{3}{*}{current}& crossing& 64.77& 65.13& 67.11& 67.40& 66.83\\ \cline{2-7}
& action& 62.77& 62.04& 74.26& 87.69& 64.39\\ \cline{2-7}
& looking& 55.64& 56.63& 38.87& 29.51& 56.94\\ \hline
\multirow{3}{*}{future}& crossing& 57.03& 61.31& 65.91& 82.82& 54.73\\ \cline{2-7}
& action& 66.46& 59.12& 79.08& 95.08& 67.69\\ \cline{2-7}
& looking& 59.08& 49.27& 20.99& 14.32& 39.26\\ \hline
\multirow{3}{*}{ \makecell{future\\skip}}& crossing& 55.08& 59.72& 64.51& 81.39& 53.42\\ \cline{2-7}

& action& 67.08& 60.48& 79.59& 96.31& 67.82\\ \cline{2-7}
& looking& 59.79& 52.20& 13.27& 8.11& 36.59\\ \hline
\multirow{3}{*}{\makecell{future\\summary}}& crossing& 44.10& 62.72& 47.34& 83.05& 33.11\\ \cline{2-7}
& action& 62.56& 58.61& 76.07& 95.08& 63.39\\ \cline{2-7}
& looking& 74.87& 49.93& 10.91& 11.54& 10.34\\ \hline
\multirow{3}{*}{\makecell{future\\skip\\summary}}& crossing& 40.51& 65.09& 47.27& 88.14& 32.30\\ \cline{2-7}
& action& 62.56& 57.09& 76.38& 96.72& 63.10\\ \cline{2-7}
& looking& 69.74& 47.95& 9.23& 11.54& 7.69\\ \hline
\end{tabular}
\caption{Evaluation results on JAAD dataset. Units are in percent.}
\label{tab:evaluate_results}
\end{table}
\subsubsection{Overall Evaluation}\label{sec:overall}
Table \ref{tab:evaluate_results} showcases our evaluation results. The term \texttt{current} represents the inferred behavior of pedestrians for given images. The terms \texttt{future} and \texttt{future skip} refer to predictions for the next five frames. These predictions are compared against two ground-truth standards: \texttt{future}, which uses a direct sequence of five consecutive frames, and \texttt{future skip}, which refers to five every 10th frames in the future. \texttt{future summary} and \texttt{future skip summary} in the table will be elaborated upon in \nameref{sec:future_summary}.

In the analysis of GPT-4V's current behavior recognition, 'crossing' shows moderate performance with 64.77\% accuracy and 65.13\% AUC, while its F1 score is a higher 67.11\%, indicating a balanced detection ability. 'action' is marked by high precision (87.69\%) but lower recall (64.39\%), suggesting accurate identification but with some missed instances. For 'looking', the model struggles more, with lower accuracy (55.64\%), AUC (56.63\%), and a significantly lower F1 score (38.87\%), indicating frequent misidentification of this behavior due to its high recall (56.94\%) but low precision (29.51\%).

In future predictions, 'crossing' sees a decline in accuracy (down to 55.08\%) and AUC (down to 59.72\%), although precision remains over 80\%, indicating correct predictions when made. 'Action' interestingly shows increased accuracy (up to 67.08\% from 62.77\%) and consistently high precision (over 95\%), suggesting improved predictive capabilities over time, but the recall remains around 67\%. 'Looking' faces decreasing performance in future predictions, with a significant drop in precision to as low as 8.11\%, pointing to challenges in accurately predicting this behavior.

Overall, GPT-4V exhibits variable accuracy and reliability across behaviors. The consistently high precision in 'action' suggests strong predictive capabilities, but lower recall across behaviors indicates a need for improvement in detecting all relevant instances. The decrease in future prediction metrics highlights the model's limitations in long-term accuracy. These varied trends across behaviors suggest a need for behavior-specific tuning to enhance GPT-4V's performance. Enhancements could include prompt adjustments or fine-tuning using diverse datasets, aiming to improve both short-term and long-term prediction reliability.
\subsubsection{Comparison with State-of-the-Art Models}
In standard pedestrian behavior prediction task, most of the works focus mainly on the crossing behavior. According to Table \ref{tab:benchmark}, Pedestrian Graph+ and PIT are leading, with the former excelling in accuracy and AUC (both 0.70), and the latter in F1 score (0.81) and recall (0.93). GPT-4V models, however, show lower accuracy and AUC scores, and their F1 scores are not at the state-of-the-art level.
While GPT-4V stands out in precision, with scores over 0.80, it falls short in recall compared to models like PIT. This high precision is vital in scenarios where false positives have serious consequences, but GPT-4V's lower accuracy and recall indicate a need for improvement, especially in identifying true positives and avoiding misclassification.
Overall, despite GPT-4V's high precision, it lags in overall performance compared to leading models, especially in accuracy and recall. This highlights the need for further development of GPT-4V to enhance its predictive accuracy and effectiveness in capturing true positives.
\begin{figure}[h]
    \centering
    \includegraphics[width=\linewidth]{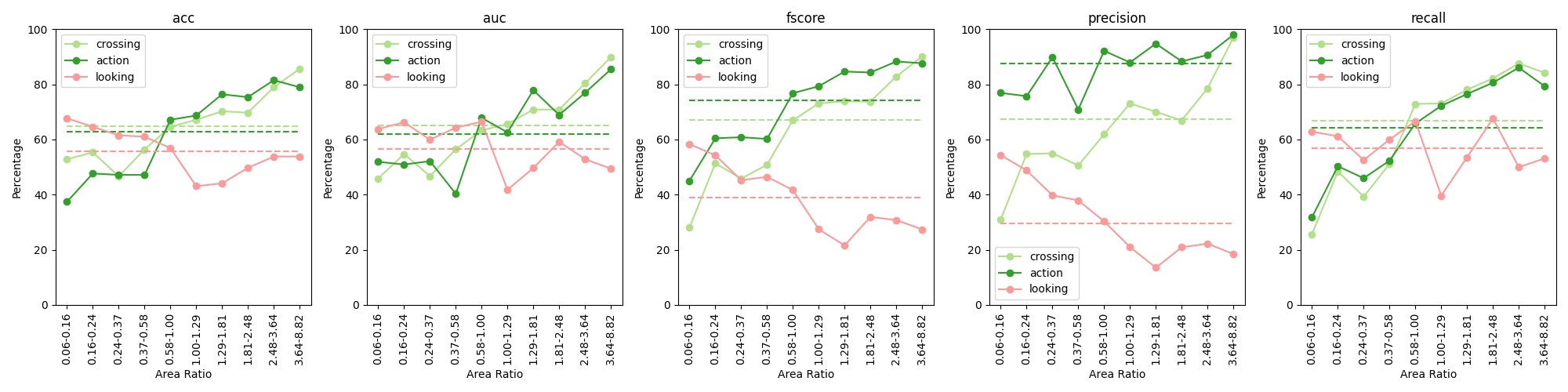}
    \caption{Accuracy of current pedestrian behavior recognition across different bounding boxes area ratios.}
    \label{fig:acc_split_area}
\end{figure}
\subsubsection{Current Pedestrian Behavior Recognition Frame by Frame}
The model struggles when pedestrians appear too small in the frames. To investigate this, we calculate pedestrian size as the ratio of the bounding box area to the total image area. We found that most of the pedestrians in JAAD dataset occupy less than 2\% of the image area.

Our analysis assesses the impact of pedestrian size on the model's capacity to discern behaviors from images. The metrics are averaged for frames within ten bins representing equal divisions of the area ratio of the bounding box. As depicted in Fig. \ref{fig:acc_split_area}, a clear pattern emerges: the accuracy increases with the size of the pedestrian, particularly in the detection of crossing and action behaviors. In particular, the accuracy exceeds 60\% for these categories once the area ratio surpasses 1.0\%. This trend suggests that the model can more easily classify larger objects. Conversely, the "looking" behavior consistently underperforms, hinting at the behaviour's intrinsic complexity or the model's limitations in recognizing this specific behavior.

\begin{figure}[h]
    \centering
    \includegraphics[width=\linewidth]{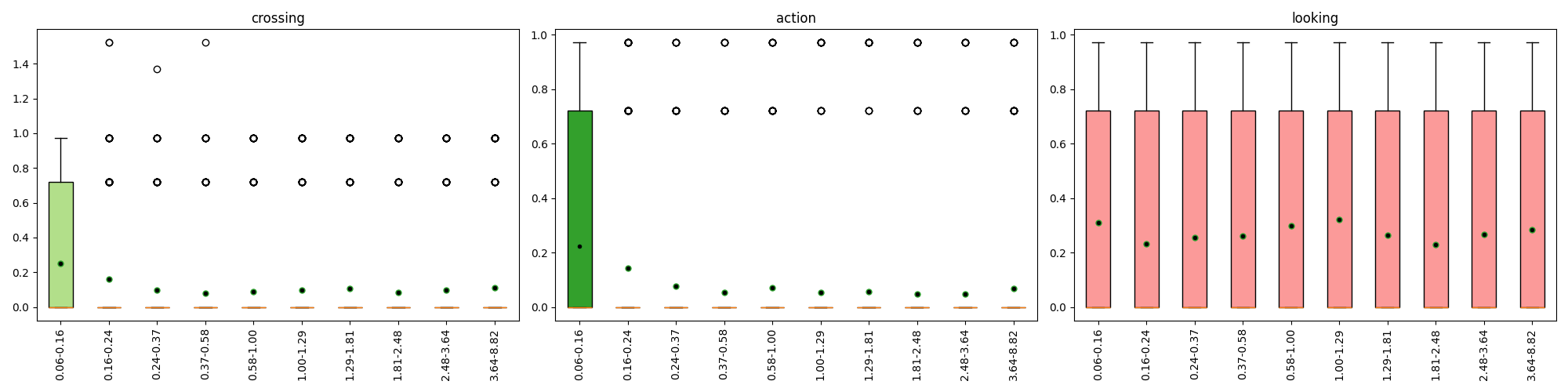}
    \caption{Average entropy of current pedestrian behavior recognition across various pedestrian size ranges.}
    \label{fig:entropy_split_area}
\end{figure}

Since GPT-4V is a generative model, we seek to understand its response variability by running five iterations per prompt and analyzing the outcomes by calculating entropy. Interestingly, the size of the pedestrians depicted in the images emerges as a factor that affects the consistency of the responses. These findings are summarized in Figure \ref{fig:entropy_split_area}. The box plots reveal that, for 'crossing' and 'action' behaviors, larger pedestrian representations lead to lower entropy, implying that predictability increases with pedestrian size. The 'action' behavior, in particular, shows greater variability at smaller sizes, as indicated by a wider interquartile range. On the other hand, the 'looking' behavior exhibits a consistently high level of entropy across all sizes. We postulate that this is due to the inherent complexity of the 'looking' task, since it relies on detecting subtle head and face orientation—a feature that occupies a very small area of the image, making it challenging regardless of the pedestrian's overall size.
\subsubsection{Future Pedestrian Behavior Prediction Frame by Frame}
This section explores the capability of GPT-4V in forecasting pedestrian behaviors over a series of time-lapsed frames. We input a set of 10 image frames into the model, each selected at every 10th interval. The model is tasked to predict the pedestrian behaviors in the subsequent five frames. The accuracy of these forecasts is measured against two benchmarks: one involving an immediate follow-up of five frames and the other considering subsequent five frames with every 10th interval. As indicated in Table \ref{tab:evaluate_results}, the prediction accuracy of GPT-4V is modest, reaching around 55\%—slightly better than a random guess. Notably, the predictions are more aligned with the continuous frame sequence rather than the skip-frame one, though the difference is minimal.
Further, we investigate how the pedestrian size in the frames influences the accuracy of future behavior predictions. Different from the analysis of current frames, we categorize our dataset into ten equally sized groups, based on the average size ratio of the input pedestrians, to maintain uniformity in the dataset.
The results show a positive correlation between the pedestrian's size in the frame and the predictive accuracy: larger pedestrian images, indicating proximity to the camera, resulting in higher accuracy and F1 scores for behaviors like 'crossing' and 'action'. However, the predictability of 'looking' behavior is more erratic. 
Additionally, we examine the consistency of the model’s predictions. It is observed that 'action' behavior predictions are generally more uniform across different groups, as indicated by lower entropy values. In contrast, the predictions for 'crossing' and 'looking' behaviors are more varied.

\subsubsection{Future Pedestrian Behavior Prediction by Summary}\label{sec:future_summary}
Because GPT-4V isn't explicitly designed to forecast pedestrian movements over several future frames, we are curious whether GPT-4V will perform better if we simplify the task by only asking about future dynamic behaviors, such as walking, crossing, and looking. Therefore, we create one ground truth label for each sequence by combining five labels. For example, if a pedestrian's future five frames are labeled as '[not-crossing, not-crossing, crossing, crossing, crossing],' we interpret this as an overall future behavior of '[crossing].' The results are thoroughly detailed in \texttt{future summary} and \texttt{future skip summary} in Table  \ref{tab:evaluate_results}.
However, summary prediction only improves certain aspects like AUC in 'crossing' and ACC in 'looking', reductions in other important metrics counterbalance this. Its effectiveness varies per behavior and does not uniformly enhance performance.
\subsection{Qualitative Analysis}
\subsubsection{Scene understanding}

In this case, we assess GPT-4V's scene understanding and its ability to prioritize pedestrians for safe navigation in a complex traffic situation. Our findings demonstrate the model's effectiveness in recognizing traffic conditions, other road users, and environmental factors. It also justifies its focus on specific pedestrians, vital for autonomous vehicle decision-making.

\begin{chatbox}[label=chat:scene_understanding,title=Scene understanding example on JAAD images]
\begin{flushright}
\begin{minipage}[t]{\linewidth}
\vspace*{0pt}
\includegraphics[width=\linewidth]{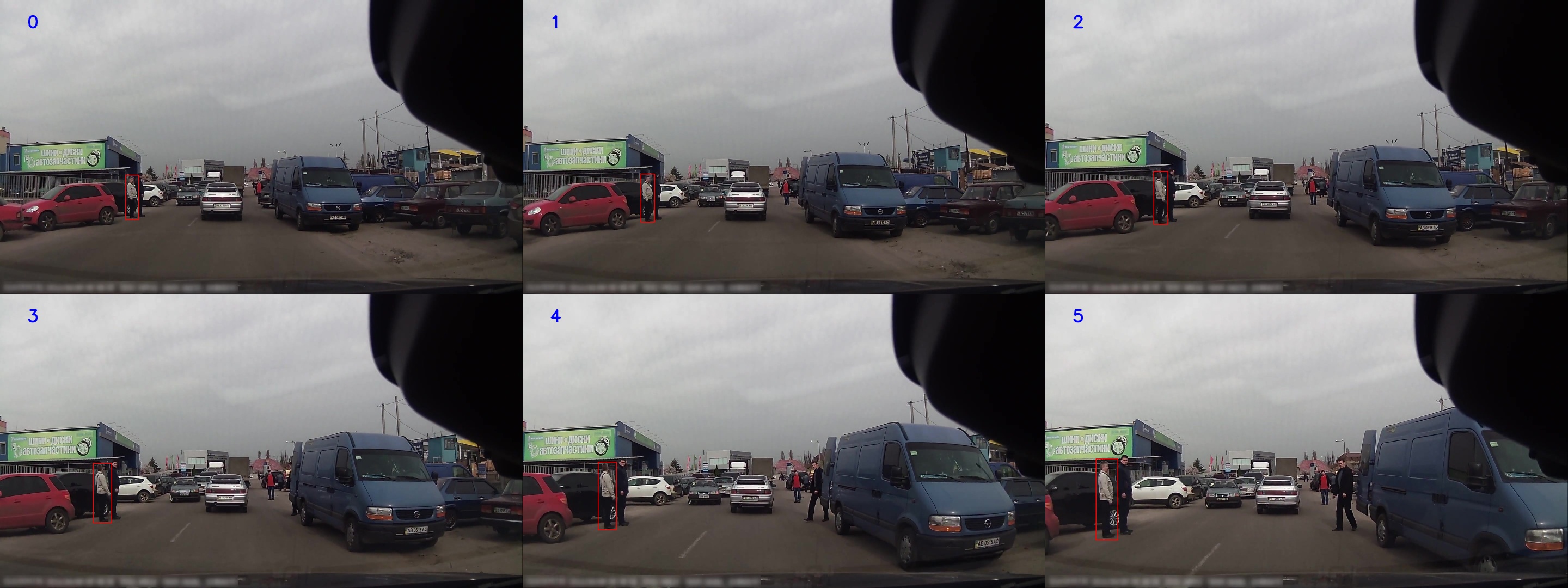}
\end{minipage}    
\end{flushright}
\begin{rightbubbles}
What can you observe in this image sequence?
\end{rightbubbles}
\begin{leftbubbles}
.....
\begin{itemize}
    \item Traffic Conditions: \textcolor{blue}{The road is busy with various vehicles}, including parked cars on the right side, indicating an active traffic situation.
    \item Pedestrian Presence: There are \textcolor{blue}{several pedestrians in the vicinity}, including at least one pedestrian in white clothing who is visible across the frames and seems to be paying attention to the road, possibly indicating an intention to cross or navigate through the parked cars.
    \item Visibility Conditions: \textcolor{blue}{The sky is overcast}, and the environment appears to be an urban or semi-urban area with commercial signage, suggesting it might be a business district or a marketplace area.
\end{itemize}
.....
\end{leftbubbles}
\begin{rightbubbles}
Among these pedestrians, who will the ego vehicle pay the most attention to to ensure safe navigation?
\end{rightbubbles}
\begin{leftbubbles}
..... The ego-vehicle should particularly pay attention to the pedestrian behind the blue car, as they are partially hidden and could potentially step into the path of the vehicle unexpectedly. .....
\end{leftbubbles}
\end{chatbox}


While GPT-4V provides a reasonably good understanding of the environment, it can struggle with subtle scene details. A case from the WiDEVIEW dataset illustrates this. The model correctly identifies colored bounding boxes and the pedestrian count within each but fails to describe their crossing behavior accurately. This is because it misinterprets a yellow-lined raised concrete as a sidewalk instead of a road section. Despite initial prompts indicating the vehicle's road presence, the model's scene understanding leads to incorrect pedestrian behavior summarizing. Additionally, since it's a single-image input, the model solely relies on scene understanding, made challenging by sunny lighting conditions.

\begin{chatbox}[label=chat:scene_understanding_failure,title=Scene understanding failure example on WideVIEW images]
\begin{flushright}
\begin{minipage}[t]{\linewidth}
\vspace*{0pt}
\includegraphics[width=0.5\linewidth]{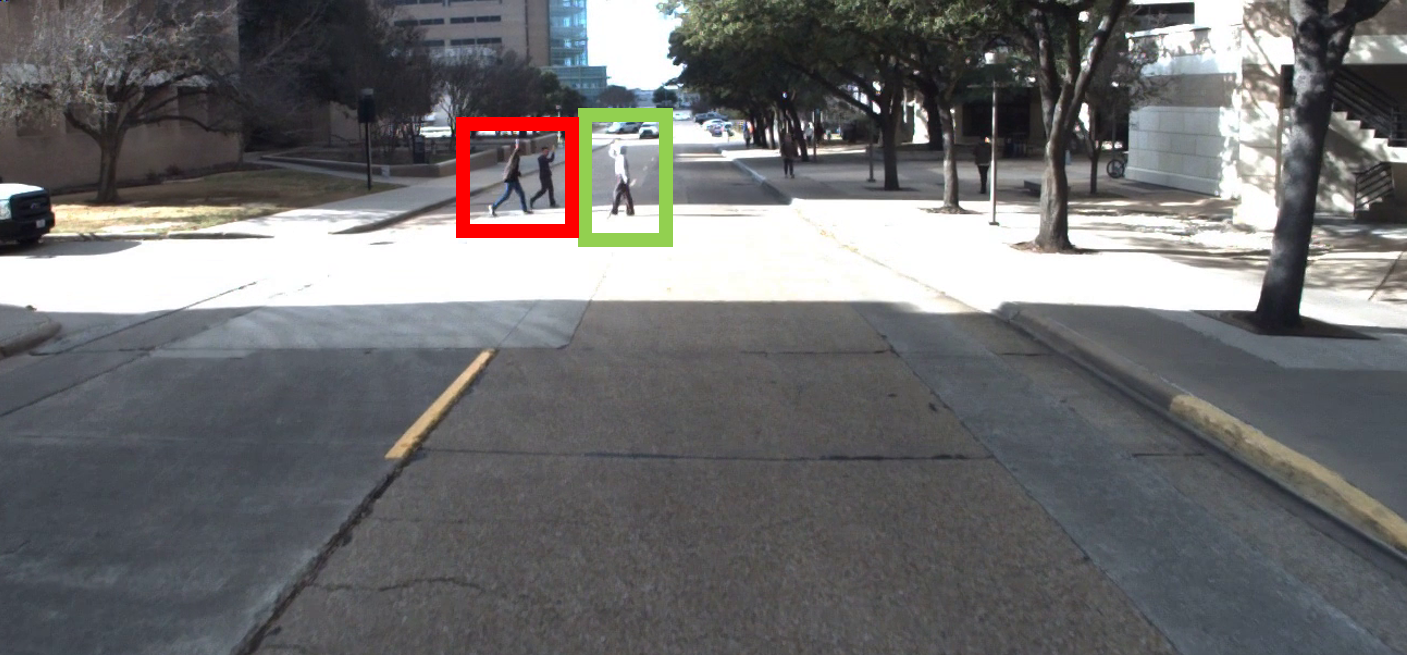}
\end{minipage}
\end{flushright}
\begin{rightbubbles}
    This is an image taken from front camera of a vehicle moving forward on the road. Can you describe this image?
\end{rightbubbles}
\begin{leftbubbles}
 The image appears to be a still shot from a vehicle's front camera, capturing a street scene during the daytime. The road is divided by a yellow line, leading to a crosswalk straight ahead. There are trees lining the sidewalk on both sides of the street, providing a sense of an urban or suburban environment. Notably, \textcolor{blue}{there are three sets of pedestrians, each highlighted with a colored rectangular box. Two are marked with a red box, and one with a green box.} \textcolor{red}{The pedestrians are walking on the sidewalk}, indicating a normal, active setting. 
\end{leftbubbles}
\begin{rightbubbles}
 Whats your reasoning behind the statement that pedestrians are walking on the sidewalk?
\end{rightbubbles}
\begin{leftbubbles}
 My reasoning is based on the visual information provided in the image:
\begin{itemize}
\item  Sidewalk Positioning: The pedestrians are within the boundaries of the sidewalk, which is separated from the road where the vehicle is likely traveling.
\end{itemize}
\end{leftbubbles}
\begin{rightbubbles}
 How is the sidewalk separated from the road where the vehicle is likely to be traveling on?
\end{rightbubbles}
\begin{leftbubbles}
 In the image, \textcolor{red}{the sidewalk is separated from the road by a raised concrete curb}. This curb acts as a physical barrier between the pedestrian space and the area designated for vehicles. Additionally, there is a grassy strip between the curb and the sidewalk, providing further separation. 
\end{leftbubbles}
\end{chatbox}

\subsubsection{Individual behavior analysis} 
In this experiment, our objective is to assess the ability of GPT-4V to identify various pedestrian or cyclist behaviors, including crossing the road and looking towards traffic, and its corresponding reasoning behind the answers.  

\textbf{Pedestrian's crossing behavior:} 
The response highlights that GPT-4V's assessment considers factors like crosswalks and intersections. When we mentioned a faded crosswalk, GPT-4V stuck to its original stance, showing a preference for certain visual cues.
\begin{chatbox}[label= chat:ped_crossing,title= Pedestrian's crossing behavior from JAAD images]
\begin{flushright}
\begin{minipage}[t]{\linewidth}
\vspace*{0pt}
\includegraphics[width=0.5\linewidth]{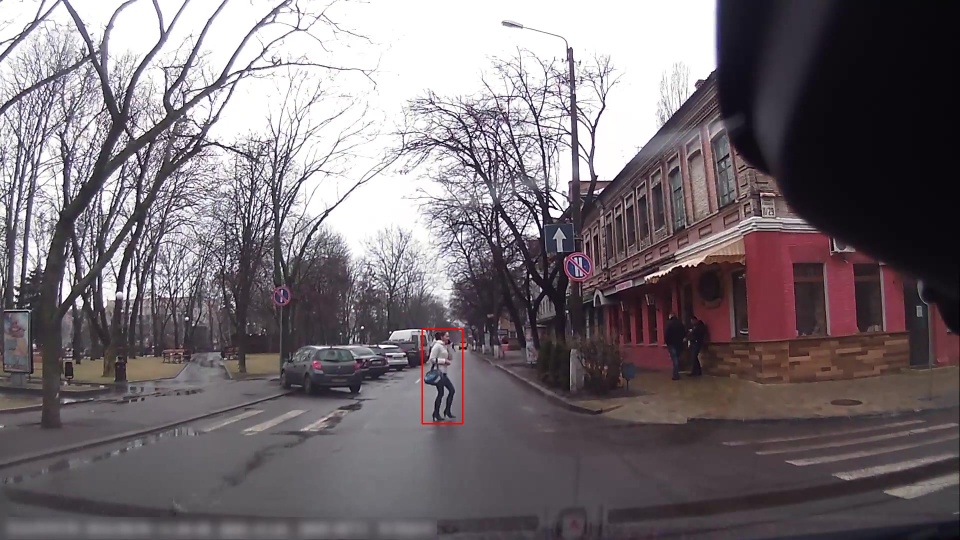}
\end{minipage}
\end{flushright}

.....
\begin{rightbubbles}
Do you think the pedestrian in the red box is trying to cross the road that the ego-vehicle is on?    
\end{rightbubbles}
\begin{leftbubbles}
...The pedestrian's body language and orientation suggest that \textcolor{red}{they are moving along the sidewalk, and there is no clear indication, like a pedestrian crossing, that they are preparing to cross the road at that exact moment.}...
\end{leftbubbles}
.....   
\begin{rightbubbles}
Is the pedestrian on a crosswalk?    
\end{rightbubbles}
\begin{leftbubbles}
...\textcolor{red}{Crosswalks are typically marked by distinct parallel lines, zebra patterns, or other road surface markings}, none of which are evident near the pedestrian in the image.    
\end{leftbubbles}
\begin{rightbubbles}
If you check the left of the pedestrian, there are some zebra patterns. It seems that the zebra patterns are worn down. 
\end{rightbubbles}
\begin{leftbubbles}
...\textcolor{red}{However, the pedestrian in the image is not directly on the patterns themselves but adjacent to them, which suggests they may be walking parallel to the road rather than actively crossing it at this moment.}...
\end{leftbubbles}
\end{chatbox}

\textbf{Pedestrian's looking behavior:} 
The model missed the looking behavior in the first and third frames, possibly due to subtle head orientation changes.
\begin{chatbox}[label=chat:ped_looking,title= Pedestrian's looking behavior from JAAD images]
\begin{flushright}
\begin{minipage}[t]{\linewidth}
\vspace*{0pt}
  \includegraphics[width=0.8\linewidth]{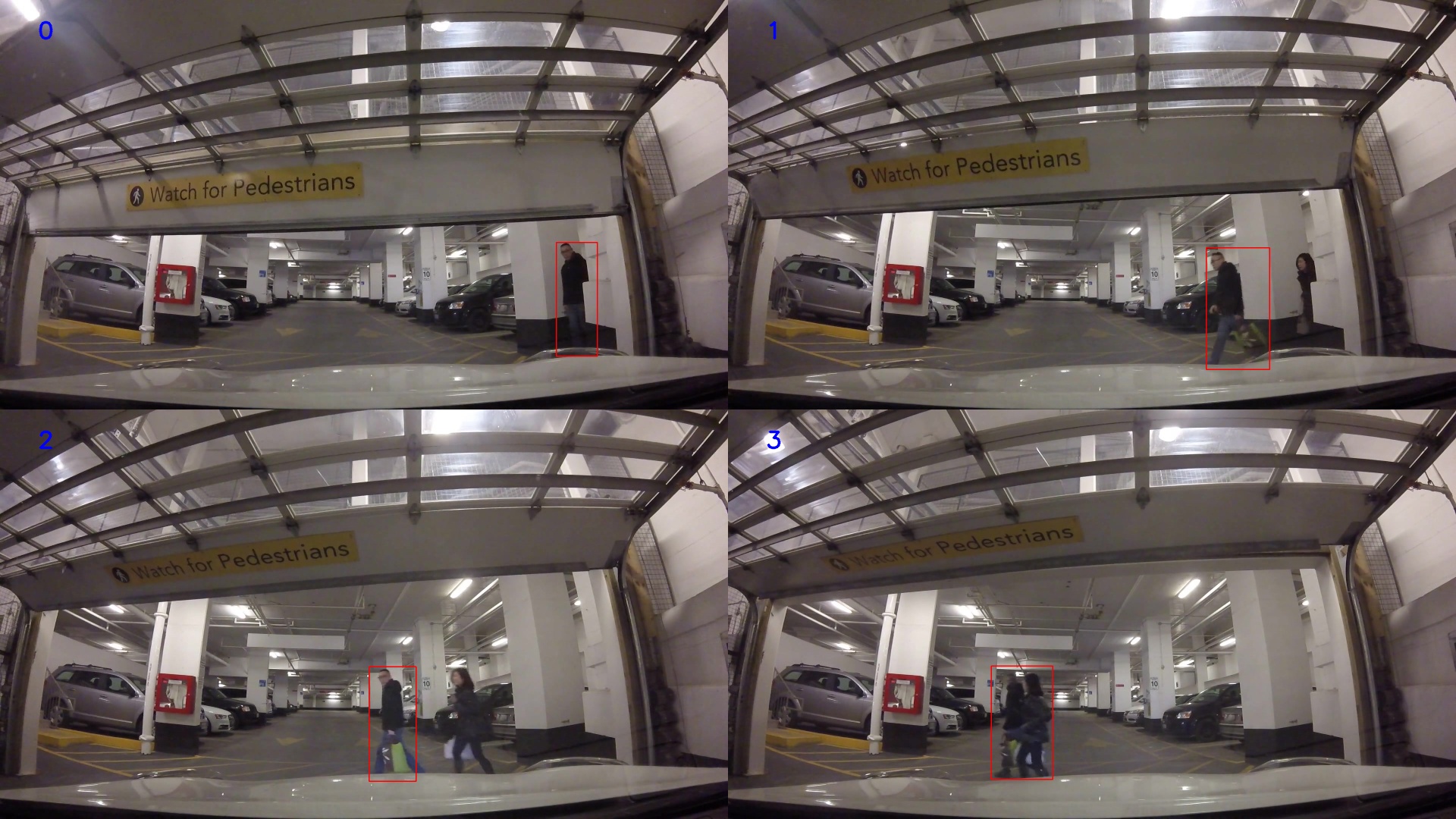}
\end{minipage}
\end{flushright}
\begin{rightbubbles}
Is the pedestrian with red rectangle looking towards the ego vehicle in this images?
\end{rightbubbles}
\begin{leftbubbles}
From the image sequence provided, the pedestrian with the red rectangle does not appear to be looking towards the ego-vehicle (the perspective of the camera). The pedestrian seems to be looking straight ahead in the direction they are walking, which is perpendicular to the direction of the vehicle. \textcolor{red}{They do not turn their head towards the camera in any of the frames}, so they would be classified as "not-looking" towards the direction of the ego-vehicle in these images.
\end{leftbubbles}
\end{chatbox}


\textbf{Cyclist's crossing behavior:} The model successfully differentiated between a cyclist and a pedestrian with a bicycle, despite the prompt mentioning only cyclists.
\begin{chatbox}[label=chat:cyc_crossing,title= Cyclist's crossing behavior from WiDEVIEW images]
\begin{flushright}
\begin{minipage}[t]{\linewidth}
\vspace*{0pt}
\includegraphics[width=0.45\linewidth]{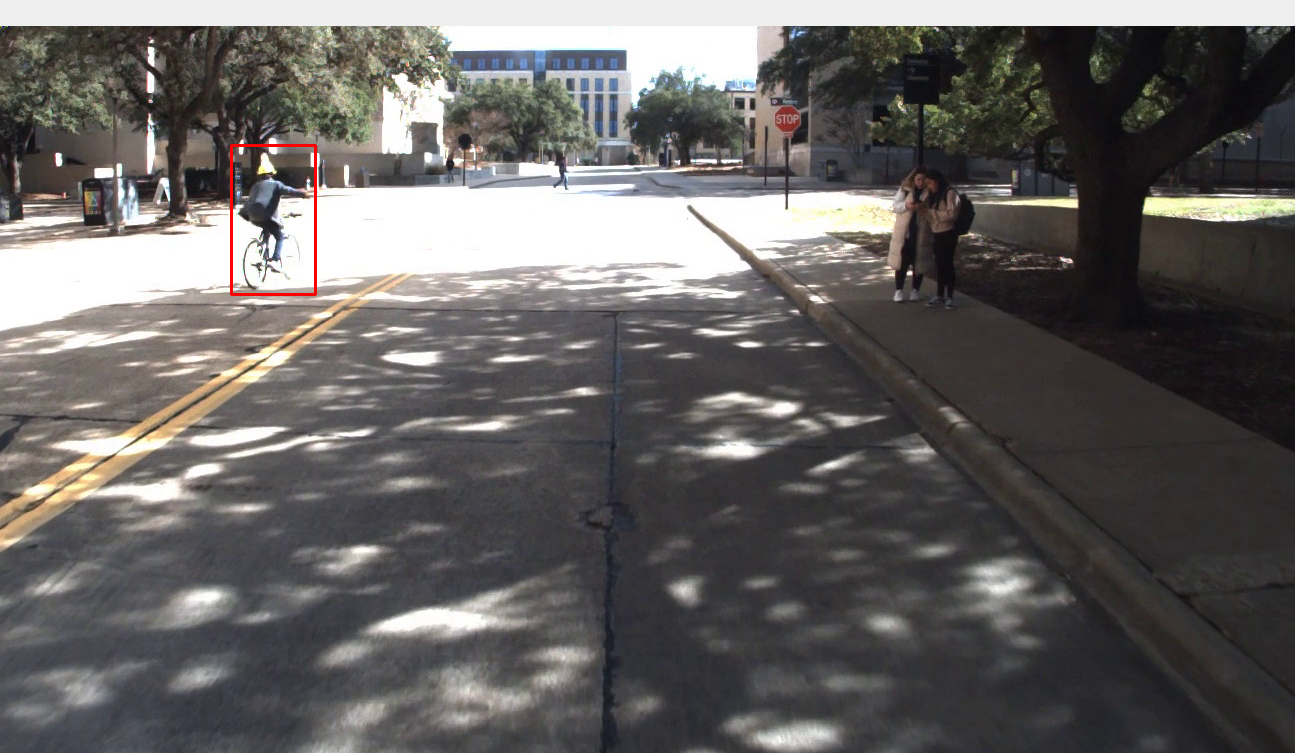}
\includegraphics[width=0.45\linewidth]{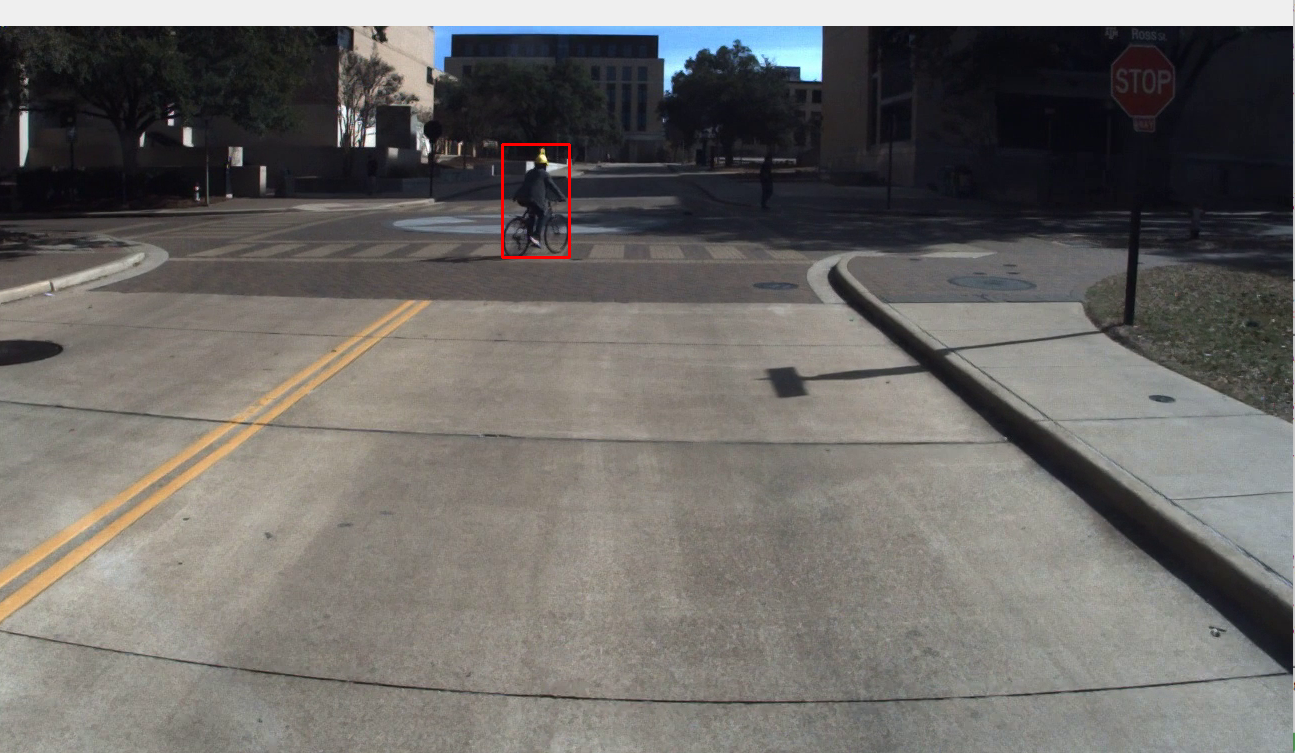}
\end{minipage}
\end{flushright}
\begin{leftbubbles}
 "summary": "The cyclist within the red box is moving across the crosswalk in a direction that intersects with the ego-vehicle's path. 
 "reasoning": "The cyclist is observed to be in motion across the frames, \textcolor{red}{moving from the right side of the road towards the left, crossing the path of the ego-vehicle.} The cyclist's head orientation suggests they are looking ahead and not towards the ego-vehicle."
\end{leftbubbles}
\begin{flushright}
\begin{minipage}[t]{\linewidth}
\vspace*{0pt}
\includegraphics[width=0.45\linewidth]{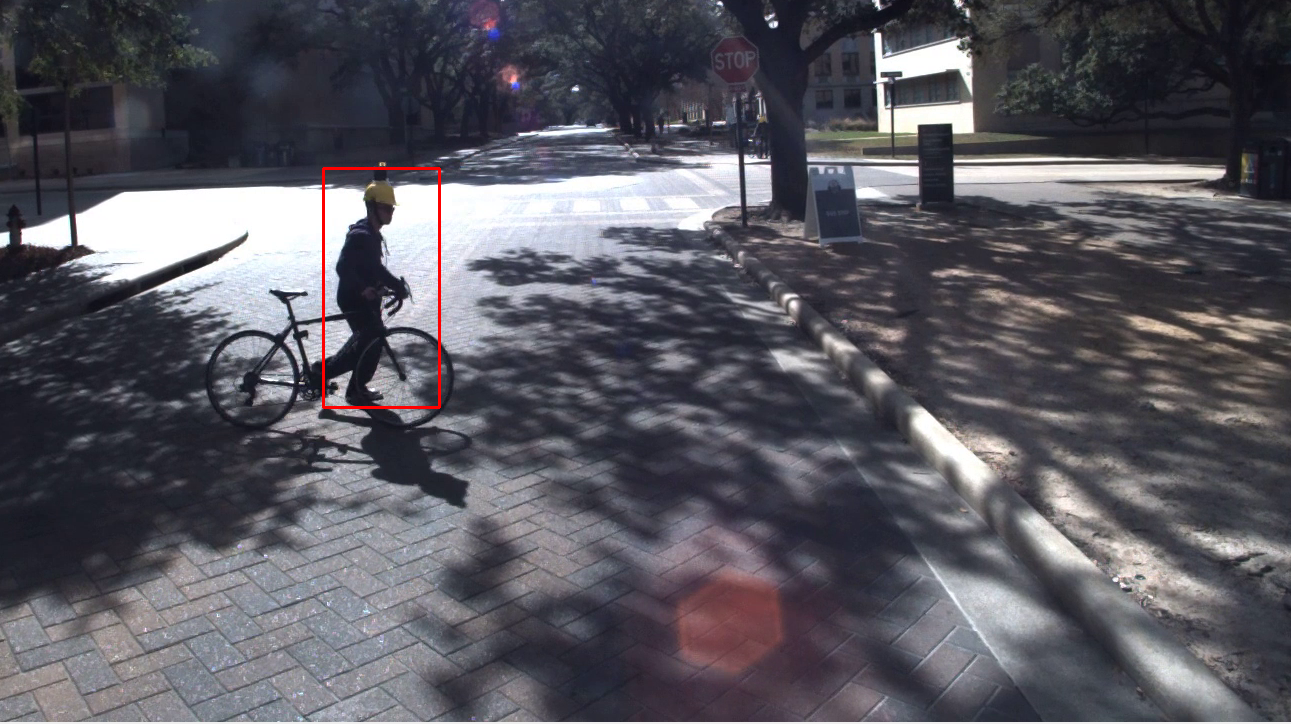}
\end{minipage}
\end{flushright}
\begin{leftbubbles}
 "summary": In the current 10 frames, \textcolor{blue}{the pedestrian within the red box is walking with a bicycle and crossing the road.} The pedestrian's path intersects with that of the ego-vehicle, and they are looking towards the direction of the ego-vehicle."
\end{leftbubbles}
\end{chatbox}
\subsubsection{Group detection and behavior analysis}



 This scenario assesses the model's ability to detect and categorize groups of pedestrians and compares its explanations to those for individual pedestrians. It's an intriguing scenario because: (1) Detecting and predicting group behavior can be highly beneficial when there's limited information about individual pedestrians. (2) Modeling group behavior can lead to better predictions, given richer historical observations. The model correctly detected the number of groups after some initial errors, but failed to describe the groups' crossing direction.

\begin{chatbox}[label=chat:group_det,title= Group detection in WiDEVIEW images]

\begin{flushright}
\begin{minipage}[t]{\linewidth}
\vspace*{0pt}
\includegraphics[width=\linewidth]{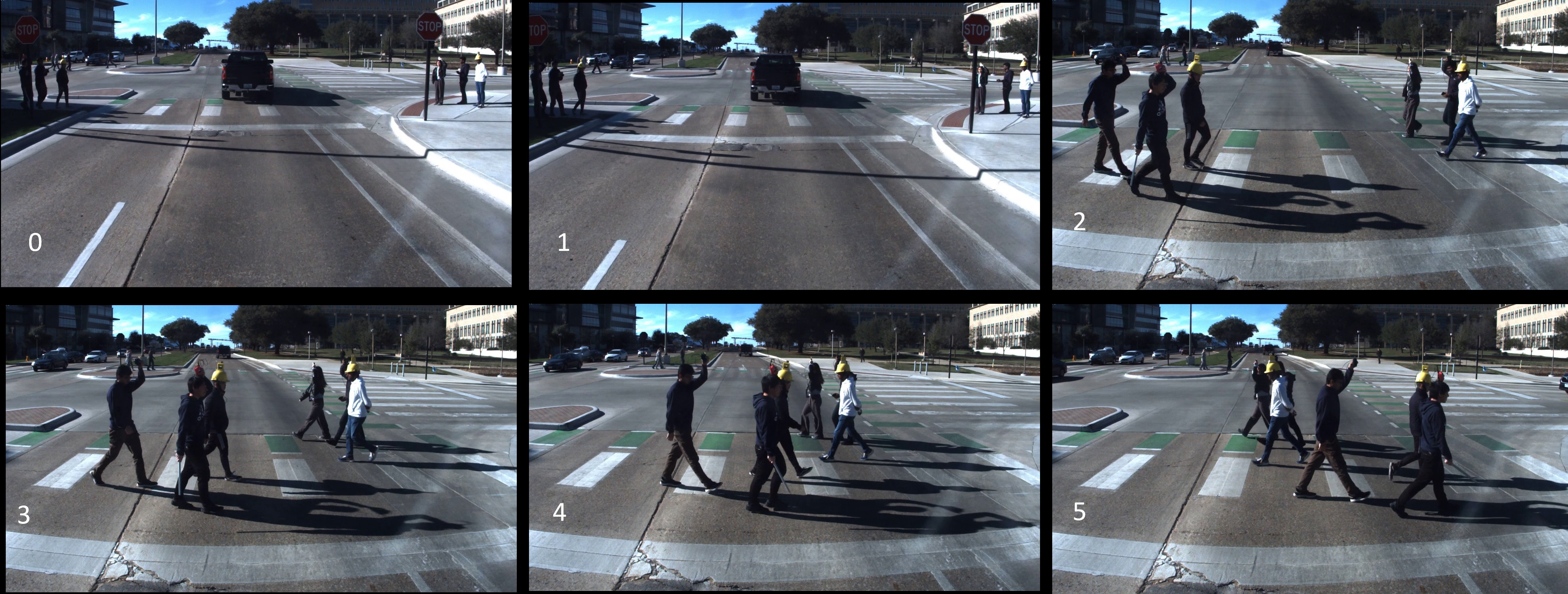}
\end{minipage}
\end{flushright}
\begin{rightbubbles}
   Can you describe the behaviours of pedestrians in this image sequence?
\end{rightbubbles}

\begin{leftbubbles}
 The image sequence appears to \textcolor{blue}{show a group of pedestrians crossing a street at a crosswalk}. The pedestrians' behavior seems orderly as they navigate the crosswalk.
\end{leftbubbles}

\begin{rightbubbles}
    Whats your reasoning behind calling the individuals a group?
\end{rightbubbles}

\begin{leftbubbles}
   The term "group" was used to refer to the collection of individuals crossing the street together. These individuals move as a loose collective, each maintaining their own pace yet moving together in the same direction across the crosswalk.
\end{leftbubbles}

\begin{rightbubbles}
   how many distinct groups of individuals do you see in the image sequence?
\end{rightbubbles}

\begin{leftbubbles}
 There are three distinct groups of individuals crossing the street:
 A pair, A trio, and A single individual in a yellow top...
\end{leftbubbles}

\begin{rightbubbles}
 how many individuals do you see walking across the street 
\end{rightbubbles}

\begin{leftbubbles}
 In the image sequence, there are five individuals walking across the street...
\end{leftbubbles}

\begin{rightbubbles}
   what about the single individual in a yellow top that you mentioned earlier?
\end{rightbubbles}

\begin{leftbubbles}
   I must correct my earlier statement. There is indeed a single individual wearing a yellow top who is part of the group of three initially mentioned. There are not two separate groups, but a single group of three individuals....
\end{leftbubbles}

\begin{rightbubbles}
   What is the direction of crossing of the pair and the trio?
\end{rightbubbles}

\begin{leftbubbles}
   \textcolor{red}{Both the pair and the trio are crossing from the right to the left relative to the camera's perspective.} As the sequence progresses, they move from the curb on the right side towards the curb on the left....
\end{leftbubbles}
\end{chatbox}
After detecting the groups, we investigate whether the model can provide additional information about group characteristics when given bounding box annotations and initial prompt information. Despite an error in pedestrian count (which was due to pedestrian being occluded through most of the frames), it accurately summarized the crossing direction and even provided insights into group formation and individual actions, like a raised arm signaling intent to cross.

\begin{chatbox}[label=chat:group_behavior,title= Group behavior in WiDEVIEW images]

\begin{flushright}
\begin{minipage}[t]{\linewidth}
\vspace*{0pt}
\includegraphics[width=\linewidth]{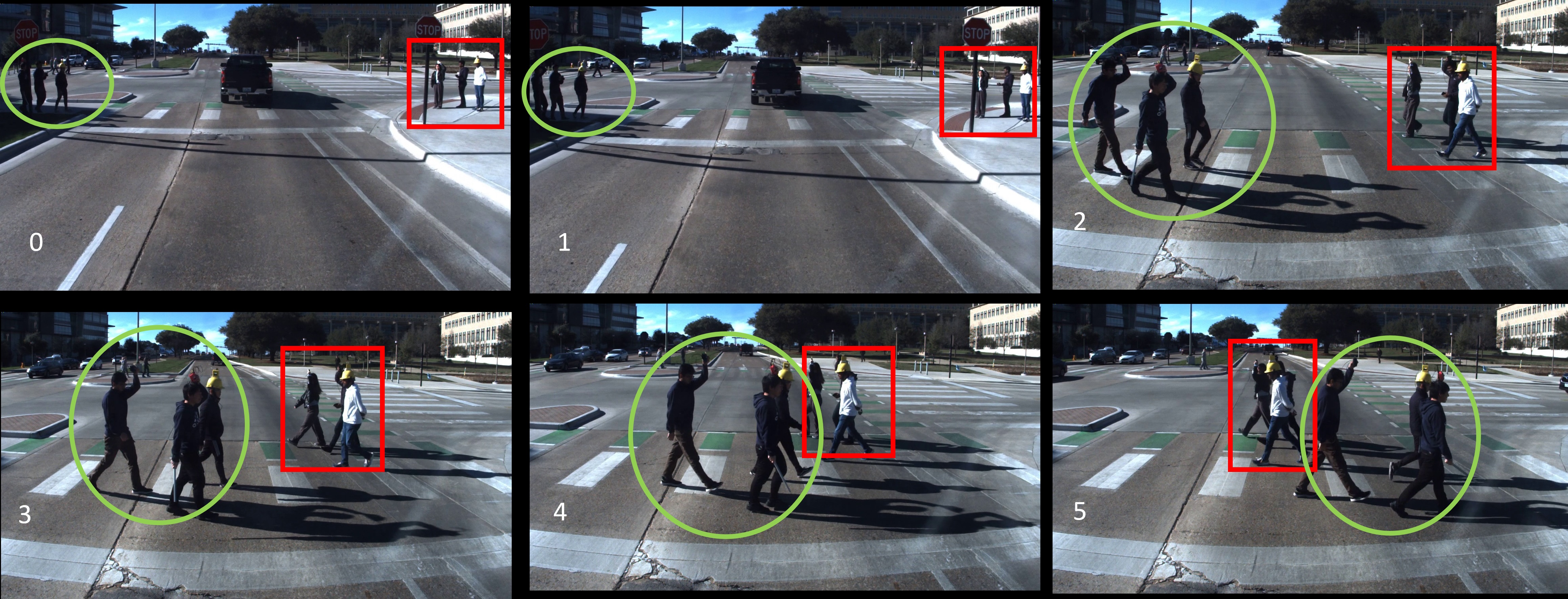}
\end{minipage}
\end{flushright}
\begin{leftbubbles}
 The group within the red box on the right side of the image appears to consist of three pedestrians. The group within the green box on the left side of the image appears to consist of two pedestrians throughout the sequence....
\end{leftbubbles}
\begin{rightbubbles}
   Are both groups crossing on the crosswalk in front of the vehicle?
\end{rightbubbles}

\begin{leftbubbles}
   Yes, both groups of pedestrians are crossing the street on the crosswalk in front of the vehicle. \textcolor{blue}{The group in the red box is crossing from the right to left of vehicle's field of view, while the group in the green box is crossing from left side to the right.}
\end{leftbubbles}
%
%
%
%
\begin{rightbubbles}
   Can you highlight the main differences in group characteristics and the crossing behaviour of each group?
\end{rightbubbles}

\begin{leftbubbles}
 Group in Red Box: This group consists of three individuals who are walking in a line, and maintain consistent spacing. Their body language is more aligned and uniform, \textcolor{blue}{indicating a coordinated group movement.}

 Group in Green Box:
Comprised of two individuals, they do not exhibit a single-file line as clearly as the group in the red box. \textcolor{blue}{One pedestrian in the green box is notably raising his arm at one point (possibly signaling his intent to the driver to cross), a distinct action not observed in the red box group. Their formation is less structured}, with the second individual walking slightly behind to the side of the first, rather than directly behind.
In summary, the group in the red box seems to have a more linear progression, while the group in the green box displays a looser formation with some individual actions (like the raised arm) that stand out.
\end{leftbubbles}
\end{chatbox}

\subsubsection{The effect of relative motion} 

This experiment demonstrates that GPT-4V struggles to recognize pedestrian behavior due to its inability to consider the relative motion between pedestrians and the ego vehicle by using an example from the JAAD dataset, where a pedestrian with a bicycle stands by the roadside as the ego vehicle approaches an intersection. Despite the pedestrian not crossing the street, GPT-4V incorrectly labels it as crossing, as observed from the viewer's perspective. This misclassification occurs because the model misinterprets the pedestrian's lateral movement, primarily caused by the vehicle's motion, as crossing, rather than the pedestrian's actual behavior. Such inaccuracies are common when pedestrians remain stationary or move parallel to the sidewalk while the vehicle's speed causes pixel position changes, leading to incorrect behavior detection.
\begin{chatbox}[label=chat:relative_motion_jaad,title= Effect of relative motion on JAAD images]
\begin{flushright}
\begin{minipage}[t]{\linewidth}
\vspace*{0pt}
  \includegraphics[width=0.8\linewidth]{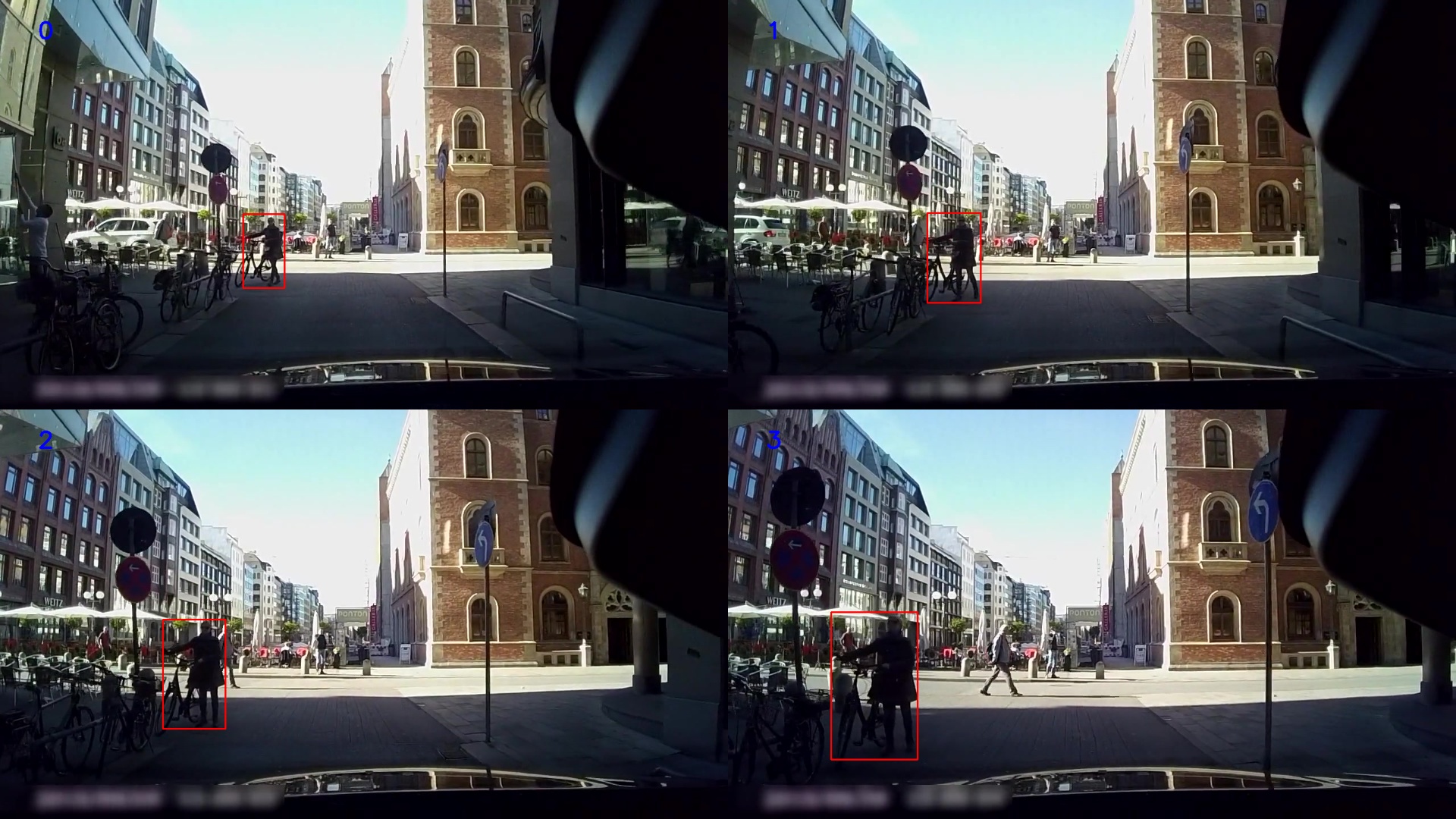}
\end{minipage}
\end{flushright}

\begin{rightbubbles}
Can you describe the pedestrians' behaviors in this image sequence?
\end{rightbubbles}

\begin{leftbubbles}
.....
\begin{itemize}
    \item In the first image, a pedestrian is about to step off the curb onto the street.
    \item The second image shows the same pedestrian taking a step on the street, seemingly beginning to cross it.
    \item In the third image, the pedestrian is further into the street, continuing to cross.
    \item The final image shows the pedestrian having crossed to the other side of the street.
\end{itemize}
\end{leftbubbles}

\begin{rightbubbles}
What's your reasoning behind this answer?  
\end{rightbubbles}

\begin{leftbubbles}
My reasoning is based on the progression of the pedestrian's position relative to the background in each frame.
.....
This \textcolor{red}{leftward movement across the frames} indicates that the pedestrian is crossing from the right to the left side of the road from the viewer's perspective.
\end{leftbubbles}

\begin{rightbubbles}
If considering the forward-moving motion of the vehicle, is the pedestrian with red box crossing or not?
\end{rightbubbles}

\begin{leftbubbles}
Considering the vehicle's forward motion, \textcolor{red}{if the relative distance between the pedestrian in the red box and the static traffic sign remains constant or similar across the images, then it is likely that the pedestrian is not actually crossing the street.} Instead, what we observe might be the result of the vehicle moving forward at a similar speed as the pedestrian walks along the sidewalk, creating the illusion that the pedestrian is moving relative to the static environment when viewed from the moving vehicle. Thus, the pedestrian with the red rectangle would not be crossing the street but walking parallel to it, maintaining a consistent pace with the vehicle.
\end{leftbubbles}
\end{chatbox}


\section{Discussion}
This section provides insights from both quantitative and qualitative evaluations, and some nuances of GPT-4V's performance, its potential, and the challenges it faces in pedestrian behavior recognition and prediction for urban driving scenarios.
\subsubsection{\uppercase{Promises}:}

  \textbf{Advanced Scene Interpretation}: GPT-4V's ability to process and interpret complex traffic scenarios is impressive. This involves not just recognizing static elements like road infrastructure but also dynamic factors like pedestrian movements and vehicle traffic flow. Moreover, the capability to prioritize the attention to perceived road users can be advantageous for safe navigation of AVs.
 
\textbf{Diverse Behavioral Understanding}: The model's ability to differentiate between various pedestrian behaviors, such as crossing, standing, and walking, mirrors human perception. This is crucial in safety-critical scenarios where recognizing subtle behavior differences can have significant consequences.

\textbf{Nuanced Differentiation Between Road Users}: GPT-4V's ability to distinguish between pedestrians and cyclists, and even between a cyclist and a pedestrian with a bicycle, shows an advanced level of detail in its analysis. This nuanced understanding is particularly beneficial in urban environments where a variety of road users share close spaces, and accurate behavior prediction for each is necessary for safe navigation.

\textbf{Group Behavior Analysis}: The model's success in detecting and analyzing groups of pedestrians is particularly promising. This aspect is not just about identifying multiple individuals but understanding group dynamics and potential collective behaviors. In crowd management and urban planning, such insights can be invaluable for designing more efficient and safer public spaces.

\subsubsection{\uppercase{Challenges}:}
\textbf{Inconsistency and Reliability Issues}:
GPT-4V faces challenges in delivering consistent and reliable outputs. Notably, it is sensitive to subtle prompt variations, which may be partly due to the deliberate introduction of randomness in the output generation process. Desired improvement in consistency was not achieved even with the temperature parameter set to minimum. This raises concerns about its stability and response reliability across different inputs.

\textbf{Dependence on Precise Prompt Structuring}: 
A well-designed prompt structure is crucial for GPT-4V's performance. We observed that consistency between prompt questions and concept definitions significantly influences the model's reasoning abilities. Higher consistency in prompt design leads to improved reasoning. Prompts that align with concept definitions is essential for coherent and reliable model responses. Additionally, designing generalized prompts for various inputs can be challenging.

\textbf{Challenges in Relative Motion Analysis}: GPT-4V’s limitations in accurately interpreting relative motion between pedestrians and vehicles can lead to critical misjudgments. This is especially concerning in scenarios where the relative speed and direction of multiple entities play a vital role in decision-making, such as in collision avoidance systems in autonomous vehicles.

\textbf{Processing Speed Limitations}: One of the limitations of the current GPT-4V version is the relatively slow processing speed, which requires 10 to 20 seconds for a single prompt used in quantitative analysis and 10 combined images. As a result, the integration of these algorithms into real-time, local systems like dashcams or traffic warning systems is currently not deemed worthwhile.

\textbf{Complex Scene Comprehension}: While GPT-4V demonstrates a good understanding of general traffic conditions, its ability to comprehend and analyze complex scenes with multiple interacting elements still requires improvement. For instance, scenarios involving simultaneous pedestrian and vehicular movements, or interactions between multiple pedestrians, present a significant challenge.

\section{Conclusion}

This study comprehensively assessed GPT-4V's capabilities to recognize and predict pedestrian behaviors in urban environment context. We carried out quantitative analysis using JAAD dataset as binary classification, and the results for crossing behavior classification were compared with state-of-the-art deep learning models. Qualitative evaluations involved interactive communication with ChatGPT4 to showcase interesting scenarios from JAAD and WiDEViEW datasets. GPT-4V exhibits promises in interpreting pedestrian behavior, particularly in diverse actions and group dynamics, with potential applicability in autonomous navigation and urban safety systems. However, challenges like output consistency, prompt structure, real-time processing, etc., need to be addressed for practical implementation in dynamic urban settings. 

\bibliography{aaai24,reference}

\begin{thebibliography}{28}
\providecommand{\natexlab}[1]{#1}

\bibitem[{Bes(2023)}]{BestPracticesPrompt}
 2023.
\newblock Best Practices for Prompt Engineering with {{OpenAI API}} | {{OpenAI Help Center}}.

\bibitem[{Chen, Tian, and Ding(2021)}]{chenGCN2021}
Chen, T.; Tian, R.; and Ding, Z. 2021.
\newblock Visual Reasoning using Graph Convolutional Networks for Predicting Pedestrian Crossing Intention.

\bibitem[{Cui et~al.(2023)Cui, Huang, Zhong, Liu, Wang, Sun, Li, Wang, and Khajepour}]{cuiDriveLLMChartingPath2023}
Cui, Y.; Huang, S.; Zhong, J.; Liu, Z.; Wang, Y.; Sun, C.; Li, B.; Wang, X.; and Khajepour, A. 2023.
\newblock {{DriveLLM}}: {{Charting The Path Toward Full Autonomous Driving}} with {{Large Language Models}}.
\newblock 1--15.

\bibitem[{Fang and L{\'o}pez(2018)}]{fang2018pedestrian}
Fang, Z.; and L{\'o}pez, A.~M. 2018.
\newblock Is the pedestrian going to cross? answering by 2d pose estimation.
\newblock In \emph{2018 IEEE intelligent vehicles symposium (IV)}, 1271--1276. IEEE.

\bibitem[{Fu et~al.(2023)Fu, Li, Wen, Dou, Cai, Shi, and Qiao}]{fuDriveHumanRethinking2023}
Fu, D.; Li, X.; Wen, L.; Dou, M.; Cai, P.; Shi, B.; and Qiao, Y. 2023.
\newblock Drive {{Like}} a {{Human}}: {{Rethinking Autonomous Driving}} with {{Large Language Models}}.
\newblock arXiv:2307.07162.

\bibitem[{Gallagher and Skalski(2023)}]{gallagherGPT4VisionComplete2023}
Gallagher, J.; and Skalski, P. 2023.
\newblock {{GPT-4}} with {{Vision}}: {{Complete Guide}} \& {{Evaluation}}.

\bibitem[{Huang et~al.(2023)Huang, Gautam, Choi, and Saripalli}]{huangWiDEVIEWUltraWideBandVision2023}
Huang, J.; Gautam, A.; Choi, J.; and Saripalli, S. 2023.
\newblock {{WiDEVIEW}}: {{An UltraWideBand}} and {{Vision Dataset}} for {{Deciphering Pedestrian-Vehicle Interactions}}.
\newblock arXiv:2309.16057.

\bibitem[{Huang, Gautam, and Saripalli(2023)}]{huangLearningPedestrianActions2023}
Huang, J.; Gautam, A.; and Saripalli, S. 2023.
\newblock Learning {{Pedestrian Actions}} to {{Ensure Safe Autonomous Driving}}.
\newblock In \emph{2023 {{IEEE Intelligent Vehicles Symposium}} ({{IV}})}, 1--8.

\bibitem[{Kotseruba, Rasouli, and Tsotsos(2021)}]{kotseruba2021benchmark}
Kotseruba, I.; Rasouli, A.; and Tsotsos, J.~K. 2021.
\newblock Benchmark for Evaluating Pedestrian Action Prediction.
\newblock In \emph{2021 IEEE Winter Conference on Applications of Computer Vision (WACV)}, 1257--1267.

\bibitem[{Liu et~al.(2020)Liu, Adeli, Cao, Lee, Shenoi, Gaidon, and Niebles}]{spatiotemporal}
Liu, B.; Adeli, E.; Cao, Z.; Lee, K.-H.; Shenoi, A.; Gaidon, A.; and Niebles, J.~C. 2020.
\newblock Spatiotemporal Relationship Reasoning for Pedestrian Intent Prediction.
\newblock \emph{IEEE Robotics and Automation Letters}, 5(2): 3485--3492.

\bibitem[{OpenAI(2023)}]{OpenAI2023GPT4TR}
OpenAI. 2023.
\newblock GPT-4 Technical Report.
\newblock \emph{ArXiv}, abs/2303.08774.

\bibitem[{Peng et~al.(2023)Peng, Wang, Dong, Hao, Huang, Ma, and Wei}]{peng2023kosmos}
Peng, Z.; Wang, W.; Dong, L.; Hao, Y.; Huang, S.; Ma, S.; and Wei, F. 2023.
\newblock Kosmos-2: Grounding Multimodal Large Language Models to the World.
\newblock \emph{arXiv preprint arXiv:2306.14824}.

\bibitem[{Quintero et~al.(2017)Quintero, Parra, Lorenzo, Fernández-Llorca, and Sotelo}]{HMM}
Quintero, R.; Parra, I.; Lorenzo, J.; Fernández-Llorca, D.; and Sotelo, M.~A. 2017.
\newblock Pedestrian intention recognition by means of a Hidden Markov Model and body language.
\newblock In \emph{2017 IEEE 20th International Conference on Intelligent Transportation Systems (ITSC)}, 1--7.

\bibitem[{Rasouli, Kotseruba, and Tsotsos(2017)}]{JAAD_cite1}
Rasouli, A.; Kotseruba, I.; and Tsotsos, J.~K. 2017.
\newblock Are {{They Going}} to {{Cross}}? {{A Benchmark Dataset}} and {{Baseline}} for {{Pedestrian Crosswalk Behavior}}.
\newblock In \emph{2017 {{IEEE International Conference}} on {{Computer Vision Workshops}} ({{ICCVW}})}, 206--213.

\bibitem[{Rasouli and Tsotsos(2020)}]{Rasouli_survey2020}
Rasouli, A.; and Tsotsos, J.~K. 2020.
\newblock Autonomous Vehicles That Interact With Pedestrians: A Survey of Theory and Practice.
\newblock \emph{IEEE Transactions on Intelligent Transportation Systems}, 21(3): 900--918.

\bibitem[{Rasouli et~al.(2022)Rasouli, Yau, Rohani, and Luo}]{Rasoulihybrid2022}
Rasouli, A.; Yau, T.; Rohani, M.; and Luo, J. 2022.
\newblock Multi-Modal Hybrid Architecture for Pedestrian Action Prediction.
\newblock In \emph{2022 IEEE Intelligent Vehicles Symposium (IV)}, 91--97.

\bibitem[{Razali, Mordan, and Alahi(2021)}]{razali2021pedestrian}
Razali, H.; Mordan, T.; and Alahi, A. 2021.
\newblock Pedestrian intention prediction: A convolutional bottom-up multi-task approach.
\newblock \emph{Transportation research part C: emerging technologies}, 130: 103259.

\bibitem[{Saleh, Hossny, and Nahavandi(2019)}]{saleh2019real}
Saleh, K.; Hossny, M.; and Nahavandi, S. 2019.
\newblock Real-time intent prediction of pedestrians for autonomous ground vehicles via spatio-temporal densenet.
\newblock In \emph{2019 International Conference on Robotics and Automation (ICRA)}, 9704--9710. IEEE.

\bibitem[{Sharma, Dhiman, and Indu(2022)}]{SHARMA2022120}
Sharma, N.; Dhiman, C.; and Indu, S. 2022.
\newblock Pedestrian Intention Prediction for Autonomous Vehicles: A Comprehensive Survey.
\newblock \emph{Neurocomputing}, 508: 120--152.

\bibitem[{Sui et~al.(2021)Sui, Zhou, Zhao, Chen, and Ni}]{sui2021tf}
Sui, Z.; Zhou, Y.; Zhao, X.; Chen, A.; and Ni, Y. 2021.
\newblock Joint Intention and Trajectory Prediction Based on Transformer.
\newblock In \emph{2021 IEEE/RSJ International Conference on Intelligent Robots and Systems (IROS)}, 7082--7088.

\bibitem[{Wen et~al.(2023)Wen, Yang, Fu, Wang, Cai, Li, Ma, Li, Xu, Shang, Zhu, Sun, Bai, Cai, Dou, Hu, and Shi}]{wenRoadGPT4VIsion2023}
Wen, L.; Yang, X.; Fu, D.; Wang, X.; Cai, P.; Li, X.; Ma, T.; Li, Y.; Xu, L.; Shang, D.; Zhu, Z.; Sun, S.; Bai, Y.; Cai, X.; Dou, M.; Hu, S.; and Shi, B. 2023.
\newblock On the {{Road}} with {{GPT-4V}}(Ision): {{Early Explorations}} of {{Visual-Language Model}} on {{Autonomous Driving}}.
\newblock arXiv:2311.05332.

\bibitem[{Xu et~al.(2023)Xu, Zhang, Xie, Zhao, Guo, Wong, Li, and Zhao}]{xu2023drivegpt4}
Xu, Z.; Zhang, Y.; Xie, E.; Zhao, Z.; Guo, Y.; Wong, K.~K.; Li, Z.; and Zhao, H. 2023.
\newblock DriveGPT4: Interpretable End-to-end Autonomous Driving via Large Language Model.
\newblock \emph{arXiv preprint arXiv:2310.01412}.

\bibitem[{Yang et~al.(2022)Yang, Zhang, Yurtsever, Redmill, and {\"O}zg{\"u}ner}]{yang2022rnnST}
Yang, D.; Zhang, H.; Yurtsever, E.; Redmill, K.~A.; and {\"O}zg{\"u}ner, {\"U}. 2022.
\newblock Predicting pedestrian crossing intention with feature fusion and spatio-temporal attention.
\newblock \emph{IEEE Transactions on Intelligent Vehicles}, 7(2): 221--230.

\bibitem[{Yau et~al.(2021)Yau, Malekmohammadi, Rasouli, Lakner, Rohani, and Luo}]{yau2021graph}
Yau, T.; Malekmohammadi, S.; Rasouli, A.; Lakner, P.; Rohani, M.; and Luo, J. 2021.
\newblock Graph-sim: A graph-based spatiotemporal interaction modelling for pedestrian action prediction.
\newblock In \emph{2021 IEEE International Conference on Robotics and Automation (ICRA)}, 8580--8586. IEEE.

\bibitem[{Zhang and Berger(2023)}]{zhang2023survey}
Zhang, C.; and Berger, C. 2023.
\newblock Pedestrian Behavior Prediction Using Deep Learning Methods for Urban Scenarios: A Review.
\newblock \emph{IEEE Transactions on Intelligent Transportation Systems}, 24(10): 10279--10301.

\bibitem[{Zhang et~al.(2021)Zhang, Abdel-Aty, Wu, and Zheng}]{pedestrian_pose}
Zhang, S.; Abdel-Aty, M.; Wu, Y.; and Zheng, O. 2021.
\newblock Pedestrian crossing intention prediction at red-light using pose estimation.
\newblock \emph{IEEE Transactions on Intelligent Transportation Systems}, 23(3): 2331--2339.

\bibitem[{Zhou et~al.(2023)Zhou, Tan, Zhong, Li, and Gou}]{zhou2023transformer}
Zhou, Y.; Tan, G.; Zhong, R.; Li, Y.; and Gou, C. 2023.
\newblock PIT: Progressive Interaction Transformer for Pedestrian Crossing Intention Prediction.
\newblock \emph{IEEE Transactions on Intelligent Transportation Systems}, 1--13.

\bibitem[{Zhu et~al.(2023)Zhu, Chen, Shen, Li, and Elhoseiny}]{zhu2023minigpt}
Zhu, D.; Chen, J.; Shen, X.; Li, X.; and Elhoseiny, M. 2023.
\newblock Minigpt-4: Enhancing vision-language understanding with advanced large language models.
\newblock \emph{arXiv preprint arXiv:2304.10592}.

\end{thebibliography}

\end{document}